\documentclass{esannV2}
\usepackage{graphicx, color, subfigure, hyperref}
\usepackage[latin1]{inputenc}
\usepackage{amssymb,amsmath,array}
\usepackage{comment}

\begin{document}
\title{TSR-DSAW: Table Structure Recognition via Deep Spatial Association of Words}

\author{Arushi Jain, Shubham Paliwal, Monika Sharma, Lovekesh Vig
%
%
\vspace{.3cm}\\
%
TCS Research, Delhi, India \\
Email:\{j.arushi, shubham.p3, monika.sharma1, lovekesh.vig\}@tcs.com
}

\maketitle

\begin{abstract}
Existing methods for Table Structure Recognition (TSR) from camera-captured or scanned documents perform poorly on complex-tables consisting of nested rows / columns, multi-line texts and missing cell data. This is because current data-driven methods work by simply training deep models on large volumes of data and fail to generalize when an unseen table structure is encountered. In this paper, we propose to train a deep network to capture the spatial associations between different word pairs present in the table image for unravelling the table structure. We present an end-to-end pipeline, named \textit{TSR-DSAW: TSR via Deep Spatial Association of Words}, which outputs a digital representation of a table image in a structured format such as HTML. Given a table image as input, the proposed method begins with the detection of all the words present in the image using a text-detection network like CRAFT which is followed by the generation of word-pairs using dynamic programming. These word-pairs are highlighted in individual images and subsequently, fed into a DenseNet-121 classifier trained to capture spatial associations such as same-row, same-column, same-cell or none. Finally, we perform post-processing on the classifier output to generate the table structure in HTML format. We evaluate our TSR-DSAW pipeline on two public table-image datasets - PubTabNet and ICDAR 2013, and demonstrate improvement over previous methods such as TableNet and DeepDeSRT.
\end{abstract}

\section{Introduction}

An important bottleneck in the digitization of camera captured or scanned documents such as invoices, resumes, and insurance forms is the presence of tabular data in an unstructured format. This tabular data is required to be digitized and stored in a machine understandable format like HTML/XML to fully capture the layout and logical structure of the table. 
Manual digitization of a large number of such document images containing tabular data is time-consuming, labour-intensive and error-prone. Hence, there is an urgent need to automate the digitization process using Table Structure Recognition (TSR) techniques. However, diversity in table layouts, complexity of nesting (spanning multiple rows and columns), multi-line texts, missing cells and overlapping columns, have made TSR a challenging task. 

Recent methods such as TableNet~\cite{TableNet}, DeepDeSRT~\cite{DeepDeSRT}, Graph Neural Networks based table recognition~\cite{TIES} often fail to work on very complex tables due to their low generalization ability. Moreover, these methods rely on very complex and deep neural network architectures which require huge amount of data and computational resources for their training. 
In addition, they ignore the implicit information available in the table images in the form of spatial association between different words which can prove to be very beneficial for capturing the table structure efficiently.
This motivates us to propose a novel and simple end-to-end pipeline for TSR which takes a table image as input and returns its structural representation in HTML format by utilizing the spatial association between different words.
Please note that we focus on extracting table structure information given an image containing table only.


The proposed method of TSR using deep spatial association of words (TSR-DSAW) is a combination of deep learning and dynamic programming techniques. It works by first detecting all the words present in the given table image using a text-detection module such as CRAFT (Character Region Awareness for Text Detection) network~\cite{CRAFT}, followed by forming relevant word-pairs efficiently using dynamic programming technique and subsequently, determining the associations between these word-pairs using a DenseNet-121 based Word2Word Association classifier e.g., same-row, same-column, same-cell or none. At the end, we generate HTML structure of the input table by applying post-processing on the outputs of word-association classifier. The proposed TSR-DSAW method is simple, robust and computationally efficient.

\section{Related Work}
\label{sec:related-work}

\textbf{Heuristics Based Methods}: Earlier conventional approaches~\cite{RuleBasedSurveyOfTableRecog, RuleBasedAutomaticTableDetection, RuleBasedExtractingTableInformationFromTextFiles, RuleBasedHeaderAndTrailerPattern} require extensive amount of human-effort for writing complex algorithms based on text-block arrangement and grid-patterns for recognizing the table layout. These methods are unable to handle the diverse layouts of the tables and the variety of grid-patterns, missing cells, row and column spans in real world datasets.

\noindent \textbf{Deep Learning Methods}: These are data-driven approaches which do not make any prior assumption about table structures and leverage deep-networks that are trained on large amount of data. One such method~\cite{DeepDeSRT} trains a deep learning network on Marmot dataset~\cite{Marmot} for locating structural components like rows, columns and cells. Later, Paliwal et al.~\cite{TableNet} proposed an end-to-end deep learning based method called TableNet for detecting table and columns by a single network where they used a single encoder and two decoders, one for each detection task. Authors in ~\cite{TableBank} demonstrate the image-to-markup model which was trained on a big TableBank dataset to predict XML from the table image directly. Though these methods work well for simple tables, they do not produce robust performance on nested tables, and tables with overlapping columns.


\noindent \textbf{Graph Based Methods}: Recently, many researchers have formulated the table recognition problem as a graph, mapping words to nodes and edges to relationships between word pairs. Qasim et al.~\cite{TIES} propose a solution with Graph Neural Networks to find the association between words on a synthetic dataset of tables. Although, they have shown remarkably good results for synthetic tables, the method does not translate well to real tables which are far more diverse in their structure patterns. In another paper~\cite{GraphBasedRAC}, authors combine the rule based approach with graph models where they propose an algorithm called RAC (Remove and Conquer) for building layout regions and then, use a graph model to describe their arrangement. Hu et al.~\cite{GraphBasedDAG} introduce a Directed Acyclic Graph (DAG) based model for hierarchical clustering to define columns and headers that are identified using spatial and lexical criteria. Very recently,~\cite{GraphBasedComplicatedTables} proposed attention based solution called GraphTSR which uses a transformer to predict the adjacency relationship of individual cells. This method made significant improvement in understanding complex tables assuming the availability of accurate bounding boxes and digital content of individual cells.

\section{Proposed Method: TSR-DSAW}
\label{sec:proposed-approach}
\begin{figure*}[h]
\centering
\includegraphics[width=\textwidth]{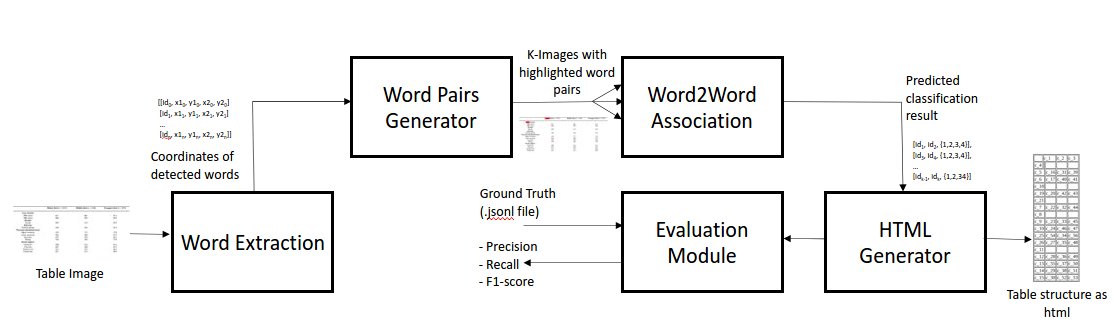}
\caption{The proposed TSR-DSAW pipeline for extracting table structure} 
\label{fig:pipeline}
\end{figure*}

TSR-DSAW takes a table image as input and comprises of $4$ modules, as shown in Figure~\ref{fig:pipeline}: (a) extraction of individual words; (b) Word-pair generation; (c) determination of Word2Word association; and (d) post-processing to generate structured output in HTML format. 

\noindent \textbf{Word Extraction}: Initially, we pre-process the input image to handle alignment issues using Radon transformation as mentioned in TableNet~\cite{TableNet}. The pre-processed image is divided into overlapping patches of size $512 \times 512$ having $50\%$ overlap with adjacent patches. These patches are then passed through CRAFT~\cite{CRAFT} which localizes individual text-regions and gives us their bounding boxes. Next, the bounding box coordinates of words present in the overlapped patches are merged which may lead to some redundancy in scenarios where the overlaps span multiple rows or columns. To circumvent this, we scan the entire image in small frames and remove the biggest box where multiple boxes cover a single frame. Finally, the extracted text-regions are read using Tesseract\footnote{Tesseract v4.1.1: https://github.com/tesseract-ocr/tesseract}.\\



\noindent \textbf{Word Pair Generation}: One trivial approach for generating word pairs is that given a word in a collection of $w$ words, make pairs with the remaining $(w-1)$ words leading to $O(w^2)$ complexity. However, this approach is inefficient. Therefore, we propose a dynamic programming based word-pair generation technique with a linear time-complexity of $O((m+n)w)$ and claim that we need a maximum of $m+n$ combinations for a word, where $m$ and $n$ are the number of rows and columns in a table. We support the correctness of our claim with the explanation that only the immediate left neighbors and top neighbors need evaluation for nested-rows and nested-columns, respectively. To handle the diversity in the size of words, we capture all the immediate words which are its top and left vicinity. We can approximate the value of $m$ and $n$ based on maximum number of text-regions to be considered in both the directions. In our experiments, we found $3$ to be an optimal value for both parameters.\\


\noindent \textbf{Word2Word Association}: This module uses a DenseNet121 based classifier to exploit the spatial relationship between different word-pairs such as 'Same Row', 'Same Column', 'Same Cell' and 'None' for identifying table-structure. In order to train the Word2Word Association classifier, we used $3600$ tables randomly selected from train-set of PubTabNet dataset to obtain input images of size $224 \times 224$ having highlighted word-pairs. The solid-red color boxes are used to represent highlighted word-pairs which also eliminates the OCR and language dependency from our model. This allows the network to entirely focus on the spatial bounds and hence, learn the spatial relationships between them. We generated $48k$ such input images covering nested/harder cases along with simpler cases. Nested cases are the word-pairs which are related with "Same Row/Column" association but structure is nested in terms of multiple-spans. On the other hand, harder cases are those pairs that are difficult to infer such as, pairs  which are quite far away but are in same cell or are near to each other but are not related in anyway. An important point to note here is that the proportion of simple and hard cases is kept equal in the training-set.\\


\noindent \textbf{HTML Generation}: This step is highly dependent on the Word2Word Association classifier results as a single wrong classification can return an entirely different structure. Assuming that spatial associations are correct, 
a directed graph is generated where nodes are cells and edge $u\rightarrow{v}$ defines $u$ and $v$ are in 'Same Row' and $u$ tends to be at left of $v$ in spatial bounds. Similarly, another graph for column information is created where edge represents $u$ to be above $v$. Graphs are traversed recursively in bottom-up fashion to collect spanning information. Row/col span of a node is defined by the sum of row/col span of its child nodes which are reachable by a single path. If a node has no child, the span is set to $1$. Using this spanning and row/column child information from the graphs, a $m\times{n}$ matrix is created, where $m$ and $n$ defines number of rows and columns respectively and the value in the matrix defines the table cell (denoted by an identifier). The resultant matrix is then converted into HTML format.

\begin{table}[!h]
  \centering
  \vspace{-2mm}
  \caption{Performance Comparison of TSR-DSAW against TableNet~\cite{TableNet} and DeepDeSRT ~\cite{DeepDeSRT}. Please note that $Acc_{w2w}$ refers to the classification accuracy of Word2Word association classifier.}
  \vspace{2mm}
  \begin{tabular}{|c|c|c|c|c|c|}
    \hline
   \textbf{Method} & \textbf{TestSet} & \textbf{$Acc_{w2w}$} & \textbf{Precision} & \textbf{Recall} & \textbf{F1-Score} \\
    \hline
    TSR-DSAW & PubTabNet  & 0.9287 & \textbf{0.9625} & \textbf{0.9087} & \textbf{0.9348} \\
    TableNet & PubTabNet  & - & 0.9572 & 0.8788 & 0.9163 \\
    \hline
    \hline
    TSR-DSAW & ICDAR13  & 0.9215 & \textbf{0.9649} & \textbf{0.9195} & \textbf{0.9416} \\
    TableNet & ICDAR13 & - & 0.9255 & 0.8994 & 0.9122 \\
    DeepDeSRT & ICDAR13  & - & 0.9593 & 0.8736 & 0.9144 \\
    \hline
  \end{tabular}
  \label{Tab:PubTabNetResults}
\end{table}

\section{Experiments}
\label{sec:exp-res}
\vspace{-2mm}
\noindent \textbf{Dataset Details}: As stated earlier, we use a subset of table images from PubTabNet train-set for training of Word2Word association classifier. For evaluation of TSR-DSAW, we select $4800$ table images randomly from val-set of PubTabNet dataset because PubTabNet has not released ground-truth for its test-set. In addition, we perform evaluation on ICDAR13~\cite{ICDAR13} which includes $152$ table images extracted from European Union and US PDF reports.\\

\vspace{-2mm}
\noindent \textbf{Evaluation Metrics}: The evaluation of table structure is done based on adjacency relations of two cell structures \cite{ProtoLinks, EvaluationMetrics}. In order to avoid the errors due to OCR, we replace the cell content and corresponding ground truth with a unique ID. We compute adjacency relations with nearest neighbors in horizontal and vertical directions. No links are created when either of the cell is blank. Next, we compute metrics~\cite{ICDAR13} such as precision, recall and f1-score in order to compare the adjacency relations between predicted and ground-truth table structure.\\

\vspace{-2mm}
\noindent \textbf{Experimental Results}: Now, we present comparison results of TSR-DSAW against two previous methods of TSR - TableNet~\cite{TableNet} and DeepDeSRT~\cite{DeepDeSRT}. It is evident from Table~\ref{Tab:PubTabNetResults} that TSR-DSAW shows significant improvement over existing methods in all the metrics on both the test datasets. For instance, in case of PubTabNet test-set, our method TSR-DSAW obtains a higher F1-score of $0.9348$ as compared to TableNet ($0.9163$). Similarly, TSR-DSAW beats both TableNet and DeepDeSRT on ICDAR 2013 test-set by a considerable margin and achieves $0.9649$, $0.9195$ and $0.9416$ as precision, recall and F1-score, respectively. An important thing to note here is that our Word2Word association classifier is trained only on images from PubTabNet train-set as mentioned in Section~\ref{sec:proposed-approach} and is evaluated on the ICDAR 2013 dataset without any further fine-tuning. Further, the classification accuracies of Word2Word association are $0.9287$ and $0.9215$ on PubTabNet and ICDAR 2013 test-sets, respectively. Due to page limit constraints, qualitative results can be found here \footnote{Qualitative Results - TSR-DSAW: https://github.com/arushijain45/TSR-DSAW}.



\vspace{-3mm}
\section{Conclusion and Future Work}
\label{sec:conclusion}
\vspace{-2mm}
We presented a deep-learning based simple method called TSR-DSAW which leverages the spatial association between words present in the table image to uncover table structure and produces an output in digital format such as HTML. The proposed method is robust in recognizing complex table structures having multi-span rows/columns and missing cells. The deep-learning based Word2Word association classifier is trained on a subset of PubTabNet train-set where it uses solid color boxes to highlight word-pairs which makes it language-independent. The proposed method is evaluated on PubTabNet validation set and ICDAR 2013 dataset without any fine-tuning. We outperform existing methods of TSR such as TableNet and DeepDeSRT on these datasets. In future, we would like to further improve the performance of Word2Word association classifier by incorporating more diverse table data and experimenting with other deep networks such as vision transformers.



\begin{footnotesize}
\bibliographystyle{unsrt}
\bibliography{mybibliography.bib}

\end{footnotesize}


\end{document}